\documentclass[11pt]{article}

\usepackage{acl}
\usepackage[T1]{fontenc}   
\usepackage{textcomp}      
\usepackage{graphicx}
\usepackage{inconsolata}
\usepackage{microtype}
\usepackage{enumitem}

\usepackage{listings}
\usepackage{xcolor} 

\lstdefinestyle{prompt}{
  basicstyle=\ttfamily\scriptsize,   
  breaklines=true,
  breakatwhitespace=false,
  columns=fullflexible,
  keepspaces=true,
  showstringspaces=false,
  frame=single,                      
  breakautoindent=false,  
  breakindent=0.5em,        
  postbreak=\mbox{\textcolor{gray}{$\hookrightarrow$}\space}, 
}

\title{Classifier-Augmented Generation\\
for Structured Workflow Prediction}

\author{Thomas Gschwind, Shramona Chakraborty, Nitin Gupta, and Sameep Mehta\\
IBM Research}

\begin{document}

\maketitle

\begin{abstract}
  ETL (Extract, Transform, Load) tools such as IBM DataStage allow users to visually assemble complex data workflows, but configuring stages and their properties remains time consuming and requires deep tool knowledge. We propose a system that translates natural language descriptions into executable workflows, automatically predicting both the structure and detailed configuration of the flow. At its core lies a Classifier-Augmented Generation (CAG) approach that combines utterance decomposition with a classifier and stage-specific few-shot prompting to produce accurate stage predictions. These stages are then connected into non-linear workflows using edge prediction, and stage properties are inferred from sub-utterance context. We compare CAG against strong single-prompt and agentic baselines, showing improved accuracy and efficiency, while substantially reducing token usage. Our architecture is modular, interpretable, and capable of end-to-end workflow generation, including robust validation steps. To our knowledge, this is the first system with a detailed evaluation across stage prediction, edge layout, and property generation for natural-language-driven ETL authoring.
\end{abstract}

\section{Introduction}

ETL and ELT (Extract, Transform, Load) workflows are foundational for data integration and analytics pipelines in modern enterprises~\cite{rahm2000data,Vassiliadis2009-survey}. These workflows combine data from disparate sources and apply structured transformations, typically authored through dedicated tools such as IBM DataStage. While such tools offer graphical interfaces for assembling workflows from predefined components, users must still manually configure transformation stages and specify dozens of low-level properties per stage making authoring tedious and error-prone even for experts.

Early work addressed these challenges through custom scripting or GUI-based simplification~\cite{vassiliadis2001arktos,reddy2010active}, and others explored semantic and ontology-driven ETL generation~\cite{jiang2010domain}. However, recent advances in large language models (LLMs) offer new opportunities to automatically synthesize such workflows directly from natural language, reducing configuration overhead and improving accessibility.

The main contribution of this paper is an end-to-end system that translates natural language utterances into executable ETL workflows.  The system is built on a \emph{Classifier-Augmented Generation (CAG)} approach, which combines utterance decomposition, classification-based stage retrieval, and few-shot prompting to predict the sequence of required workflow stages. These predicted stages are then assembled into a directed flow via edge prediction and configured via property prediction based on stage-local sub-utterances.

Our architecture is modular, interpretable, and supports robust validation. We compare CAG against strong single-prompt and agentic baselines, demonstrating improvements in both accuracy and efficiency (e.g., over 60\% token reduction while at the same time using a smaller model). Our system is already integrated into a production ETL tool (IBM DataStage), where it supports real-world user workflows, serving both novice users, who benefit from reduced interaction complexity, and expert users, who gain from auto-filled configurations requiring only a lightweight review.

We report detailed evaluation results for each generation step (stage, edge, and property prediction) and discuss how system design decisions impact performance and interpretability. Our findings provide practical guidance for building LLM-assisted authoring tools for structured, real-world automation tasks.

\section{Background}

We selected IBM DataStage due to its scale, mature ecosystem, and publicly documented stages and configuration interfaces~\cite{ibm-datastage-docs}, making it a representative platform for evaluating structured workflow generation. DataStage enables users to design and run ETL (Extract, Transform, Load) and ELT workflows across diverse data sources using a visual interface, where predefined components (e.g., connectors, transformers, and aggregators) are assembled into workflows.  

As a one-time setup, we extracted 142 DataStage stages (90 of them datasource connectors) along with their descriptions and properties from the official documentation. Stages, also referred to as datasources, operators, or tasks, form the basic building blocks of a workflow. Each stage has between 1 and 111 properties, with an average of~27.6. For every property, we store its description, type, default value, and availability conditions in a structured format suitable for prompt generation and output validation.


\section{Workflow Generation}

This section introduces and evaluates our modular pipeline for generating ETL workflows from natural language. We begin by comparing different approaches for predicting the set of workflow stages (Sections~\ref{sec:single}--\ref{sec:cag}), then use the best-performing approach to predict edges between these stages (Section~\ref{sec:edges}), and finally generate stage properties (Section~\ref{sec:properties}). For this evaluation, we sampled 1010 natural language flow descriptions for stage prediction. \footnote{These utterances are paraphrased and anonymized from internal real-world DataStage flows, including customer-inspired examples. No IBM documentation was used to construct these flows.}  From this set, we annotated 308 flows with 1410 properties, and additionally sampled 54 complex non-linear flows (with up to 14 stages) for edge prediction.

\subsection{Single Prompt Stage Prediction}
\label{sec:single}

The first approach uses a single prompt that presents all 142 stages and asks the LLM to identify those needed for the workflow. The prompt includes task instructions, stage names with one-line descriptions, 142 few-shot examples, and the user’s utterance. Few-shot examples help compensate for the LLM’s lack of pretrained knowledge about individual stages~\cite{Kojima2022llms}, showing how stages are combined in real tasks. On average, each stage appears in about two examples.





As shown in Table~\ref{tab:acc_single}, the LLMs perform well, especially the bigger ones, in predicting the workflows at hand.  In certain cases, the models appear to overlook relevant stages or parts of the utterance, likely due to the sheer number of stages presented in the prompt.  For instance, given the utterance ``Combine the employee\_info master dataset with the employee\_updates and department\_changes datasets on employee\_id. Once done, update the employee\_records and employee\_department information accordingly.'' most models only returned the \emph{join\_merge} instead of the correct \emph{join\_merge} and \emph{modify} stages. The LLaMA3.2 model only returns \emph{combine\_records}, which in combination with its poor accuracy shows that this task exceeds the model's capabilities.  As explained in~\cite{brown2020-language-models-are-few-shot-learners}, larger LLaMA models outperform smaller models.

\begin{table}[tb]
    \centering
    \small
    \begin{tabular}{l|rrr}
          & \multicolumn{3}{c}{Accuracy [\%]} \\
        Model                  & total & $1$-op & $n$-op \\
        \hline
        llama-3.2-3b-instruct    &  71.1 &   92.3 &   23.2 \\
        granite-3.1-8b-instruct  &  88.0 &   92.6 &   77.7 \\
        llama-3.3-70b-instruct   &  96.4 &   98.1 &   92.6 \\
        llama-4-mvk-17b-128e-\ldots\footnotemark & 95.8 & 97.7 & 91.6 \\
    \end{tabular}
    \caption{Single-prompt accuracy on 1010 flows; "$1$-op" refers to predictions involving a single stage; "$n$-op" refers to predictions involving two or more stages}
    \label{tab:acc_single}
\end{table}

\footnotetext[2]{llama-4-maverick-17b-128e-instruct-fp8}

Because this approach includes all the DataStage stages and many few-shot examples in the prompt, it sends almost $14'000$~tokens per request to the LLM.  The variation in tokens used for the different requests is relatively small (+/-100) since the prompt, except for the user's utterance, remains the same for every request.  Despite the large prompt size, modern LLMs with 128k context windows~\cite{grattafiori2024-llama-3-herd-models, stallone2024-scaling-granite-code-models} can process such inputs comfortably.





\subsection{Agentic Stage Prediction}
\label{sec:agentic}

To reduce prompt size and improve modularity, we adopt an agentic approach in which the LLM operates as a ReAct-style agent~\cite{yao2023react}. Rather than receiving all stage definitions up front, the LLM itself decomposes the user utterance into sub-utterances and invokes a callable stage classification tool to identify appropriate components. This setup is inspired by recent work on integrating external tools into LLM workflows~\cite{Schick2023-toolformer, Qin2024tool, Qu2025tool}.

The classification tool uses a fast and small classification model trained on a set of $2'697$ (utterance, operator) single-label pairs covering 138 semantic labels (derived by merging close variants among the 142 total stages).\footnote{The dataset is split into 2'133 training examples and 564 test samples.  Evaluation yields macro-Precision of 98.6\%, macro-Recall of 97.8\%, and macro-F1 of 98.1\%, based on sigmoid-thresholded outputs across classes.}  Since our training data consisted only of single-label pairs, we could not train a multi-label classification model.  As base models, we used once Meta AI's RoBERTa-large model~\cite{liu2019roberta} and once IBM's slate-125m-english-rtrvr model~\cite{ibm_slate125m_rtrvr}, both are small enough and run extremely fast even on a standard CPU.  After training, both models had an accuracy of about 98\% with RoBERTa-large performing marginally better but also a bit slower.  Since speed is of importance to use, we use slate as our base model.

Since the classification model operates on single-label inputs, we expose it to the LLM as a callable tool.  The prompt itself contains instructions explaining the LLM's task, the available classification tool, and the user's utterance.  The LLM correctly splits the utterance into smaller sub-utterances that each describe a stage to be executed and passes those sub-utterances to the classification tool.  Based on the results of the classification tool, the LLM generates the final answer for the flow.

When examining the results in Table~\ref{tab:acc_agentic}, we observe that LLaMA4 underperforms compared to LLaMA3.3, despite dedicated prompt tuning. Similar weaknesses of LLaMA4 on specific tasks have also been reported by others\cite{nuenki_llama4_translation, reddit_llama3_gt_llama4}. Overall, the agentic approach performs worse than the single-prompt method across all models. We identify three main reasons: (1)~the \emph{granularity mismatch} between generated sub-utterances and the functional scope of DataStage stages (some sub-utterances are too fine-grained, others too coarse); (2)~\emph{semantic similarity} between certain stages, such as \texttt{split\_subrecord} and \texttt{split\_vector}, which the limited-capacity classifier struggles to disambiguate; and (3)~occasional \emph{classification failures}, even for simple cases, which, though infrequent, can reduce user trust in the system.

\begin{table}[tb]
    \centering
    \small
    \begin{tabular}{l|rrr}
          & \multicolumn{3}{c}{Accuracy [\%]} \\
        Model                  & total & $1$-op & $n$-op \\
        \hline
llama-3-2-3b-instruct          &  33.4 &   40.0 & 18.4   \\
granite-3-1-8b-instruct        &  45.6 &   37.0 & 65.2   \\
llama-3-3-70b-instruct         &  69.3 &   90.4 & 21.6   \\
llama-4-mvk-17b-128e-\ldots    &  40.0 &   27.9 & 67.4   \\
\end{tabular}
    \caption{Agentic accuracy (1010 flows)}
    \label{tab:acc_agentic}
\end{table}


The challenge of this approach is that without any additional information, it is very hard for the LLM to split the user's utterance into sub-utterances of the right granularity. For instance, the best performing LLaMA 70b model, for the utterance ``I want to use teradata where my connection name is teradata-00, schema name is TM\_DS\_DB\_1 and table name is EMPLOYEE2. then sort on the age column. then filter out pizza column. then postgres where my connection name is tristan\_postconn , schema name is public and table name is demoautotest, Also do the following, Decimal rounding mode is ceiling, Generate Unicode Columns, Row limit should be 50.'', formed the sub-utterance (``sort on the age column, filter out pizza column, Decimal rounding mode is ceiling, Generate Unicode Columns, Row limit should be 50'')  containing multiple stages which the classification tool classifies as \emph{sort}.  The model realizes that this cannot be complete and in subsequently asks the classification tool to classify ``filter out pizza column'' (\emph{filter}), ``Decimal rounding mode is ceiling'' (\emph{decode}), ``Generate Unicode Columns'' (\emph{column\_generator}), and ``Row limit should be 50'' (no match).  Unfortunately, while trying to figure all this out, the model omits the seemingly trivial upstream and downstream database stages (\emph{teradata} and the \emph{postgresql}) likely due to context loss or attention limits.

We tried to mitigate this by providing a set of examples of how to split an utterance into sub-utterances to use for the classification tool.  Except, this confused the LLMs and they interpreted these examples as input-output examples with the result that the output of the LLM became mostly unusable.  Considering that the LLMs were able to do some consolidation of the final results and did not simply return the concatenations of the classification results, we tried to return a description of the stage predicted by the classification function clearly marked as stage description but that too got misinterpreted by the LLMs and again the accuracy dropped considerably.  Finally we tried to provide the LLM with a function that instead of giving a single recommendation, simply suggests candidates and documented this as part of the tool description but again the LLM misinterpreted the output and thought that all of these tasks should be included in the final flow.


\subsection{Classifier Augmented Stage Prediction}
\label{sec:cag}
\label{sec:classification}

Agentic approaches enable language models to autonomously reason and act, but in our experiments they struggled with utterance decomposition, tool invocation, and prompt interpretation.  As in the agentic approach, we ask the LLM to split up the user's utterance into single-stage sub-utterances which we then classify with our classifier tool.  The stages identified through these tool calls form a set of candidate stages. These are then passed to the LLM along with one-line descriptions and few-shot examples that specifically support them, hence the name Classifier Augmented Generation (CAG).  CAG addresses practical shortcomings of the agentic setup while retaining compatibility with agent-based workflows.

This approach has several other key benefits.  The generation of the sub-utterances is a separate step in which we can provide few-shot examples demonstrating utterances and the corresponding sub-utterances.  While these examples may not provide a complete overview of the granularity of every stage, they do give the model already a rough idea of the expected granularity.  Additionally, after the classification model has identified the candidate stages, we scan the user's utterance for stage names (e.g., filter) and their synonyms (e.g., extract) and if they match add the corresponding stages to the set of candidate stages.  This keyword search significantly reduces mispredictions for seemingly trivial utterances.

Once the set of candidate stages has been computed, we use the prompt from Section~\ref{sec:single} limited to the stages identified as candidates and about 40 few-shot examples that mention at least one of the stages in the candidate set.  This way, we use both the classifier and keyword search to narrow the candidate stage space, while the LLM performs the final multi-label prediction. Additionally, this approach allows us to present a larger number of few-shot examples relevant for candidate stages, enabling the LLM to better distinguish between similar stages.

\begin{table}[tb]
    \centering
    \small
    \begin{tabular}{l|rrr}
          & \multicolumn{3}{c}{Accuracy [\%]} \\
        Model                  & total & $1$-op & $n$-op \\
        \hline
        llama-3.2-3b-instruct  &  90.1 &   97.6 &   73.2 \\
        granite-3.1-8b-instruct  &  94.0 &   97.9 &   85.2 \\
        llama-3.3-70b-instruct &  97.2 &   98.6 &   94.2 \\
        llama-4-mvk-17b-128e-\ldots & 97.7 & 99.0 & 94.8 \\
    \end{tabular}
    \caption{CAG accuracy (1010 flows)}
    \label{tab:acc_hybrid}
\end{table}

Using this approach, we arrive at the accuracy numbers shown in Table~\ref{tab:acc_hybrid}, consistently performing better than the already excellent single prompt approach.  Moreover, this improved performance comes at a lower cost: the approach uses only 4,000–4,700 tokens per request on average, depending on the model, reducing the number of tokens per request by approximately~66\%.  In addition, it achieves comparable accuracy while using a smaller model that is significantly cheaper per token, resulting in further overall cost savings.
As in other settings, we observe accuracy increasing with model size.
One illustrative failure case is the utterance: ``Split the full\_name field of the employee\_data dataset into separate columns for first\_name and last\_name, then capitalize the first letter of each name for consistency.'' Most models return only the \emph{split\_subrecord} stage and omit the expected \emph{modify} stage.  Although the models correctly identify ``Capitalize the first letter of each name for consistency'' as a distinct sub-utterance, our classifier incorrectly maps it to \emph{head} stage.  Because the keyword search also fails to surface the \emph{modify} stage, it is excluded from the prompt and thus not predicted.  The model correctly rejects \emph{head} based on the few-shot examples, but with no other relevant candidate, it defaults to predicting only \emph{split\_subrecord}.


\subsection{Edge Predictions}
\label{sec:edges}
\label{sec:nonlinear}

So far, the stage and property prediction components operate under the assumption of linear flows, where the execution order is implicitly defined by the sequence of stages. However, many real-world ETL workflows are non-linear: they include branching, parallel processing, joins, and multiple input or output sinks. In these cases, the data flow structure cannot be inferred from the stage order alone, and edges must be explicitly predicted.

We define non-linear flows as workflows involving branching, joining, or parallel transformations that cannot be expressed as a single linear pipeline.  We generate the flow structure based on the stages predicted by the CAG approach (95\% accuracy).

First, to distinguish repeated stages, we assign unique names (e.g., by adding a ‘\_$n$’ suffix) while preserving their semantic identity.  Second, to anchor these names to a specific part of the user's utterance, we ask the LLM to split the utterance according to the stages that we have obtained through the classification approach.  Since the LLM now has access to the task list and stage descriptions, it can segment the user's utterance into sub-utterances with over 99\% accuracy.  The result is a node list, where each node includes a unique name, a localized task description, and cardinality constraints derived from DataStage specifications. Given this input and the original utterance, the LLM predicts the flow structure.

Sometimes, the predicted edges violate the predefined input/output cardinality constraints of certain stages, particularly when the CAG step fails to identify all necessary stages.  This can lead to constraint violations.  To address this, we validate whether each node's edge count matches its allowed cardinality. If violations are found, we attempt to split overloaded nodes (e.g., a node with no inputs and multiple outputs), if such a split is unambiguous. If splitting is not possible, we remove excess edges until all constraints are satisfied.

\begin{table}[tb]
    \centering
    \small
    \begin{tabular}{l|rrr}
        Model                       & similarity & exact \\
        \hline
        llama-3-2-3b-instruct       &       31\% &   0\% \\
        granite-3.3-8b-instruct     &       41\% &   4\% \\
        llama-3.3-70b-instruct      &       73\% &  37\% \\
        llama-4-mvk-17b-128e-\ldots &       42\% &  15\% \\
    \end{tabular}
    \caption{Edge prediction (54 non-linear flows)}
    \label{tab:nonlinear}
\end{table}

Table~\ref{tab:nonlinear} shows the results for complex non-linear flows with 6--14~stages (avg~8.1) per flow (again, LLaMA4 struggles with some tasks), it is the most challenging step, even the best model achieves only 37\% exact match and 73\% structural similarity. These results are still valuable as flows with 70–80\% similarity often require only minor corrections.  For this aspect we have not yet managed to optimize the prompts, so there may be room for improvements.

\subsection{Property Prediction}
\label{sec:properties}

To predict stage properties, we use the sub-utterances generated for each individual stage by the edge prediction.  Using the full utterance can lead to ambiguity, as LLMs may struggle to determine which property belongs to which stage—especially when a stage appears multiple times (e.g., multiple \emph{sort} stages sorting different data artifacts).  Hence, properties are predicted individually for each stage.  Each prompt contains task-specific instructions, the sub-utterance, the corresponding stage name, a list of all supported properties (each with a one-line description), and a one-shot example.

\begin{table}[b]
    \centering
    \small
    \begin{tabular}{l|rrr}
        Model                       & prec. & recall &   F1 \\
        \hline
        llama-3-2-3b-instruct       &  88\% &   72\% & 0.79 \\
        granite-3.3-8b-instruct     &  93\% &   71\% & 0.81 \\
        llama-3.3-70b-instruct      &  92\% &   81\% & 0.86 \\
        llama-4-mvk-17b-128e-\ldots &  94\% &   67\% & 0.78 \\
    \end{tabular}
    \caption{Property prediction (308 flows with 1410 properties)}
    \label{tab:property_prediction}
\end{table}

\begin{figure*}[t]
    \centering
    \includegraphics[scale=0.75,       
        trim=0.6cm 14.2cm 8.5cm 1.05cm,   
        clip                   
]{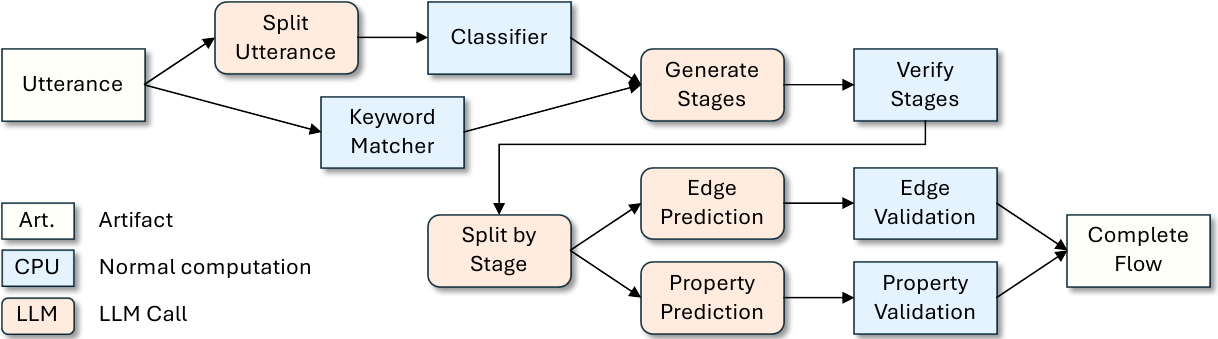}
    \caption{Interaction between CAG, edge prediction, and property prediction to generate workflows.}
    \label{fig:architecture}
\end{figure*}

To mitigate errors in generated properties, we apply a multi-dimensional validation strategy that validates the properties associated with the current stage.  First, we remove any generated properties whose names do not match a valid property defined for that stage.  Second, we check whether the value of each property can be coerced to the property's declared type. If type coercion fails, the property is discarded.  Third, we evaluate inter-property dependencies: some properties are only valid if others are present with specific values. This logic is encoded as metadata expressions evaluated at runtime, with access to the full set of properties predicted for the stage. For instance, in the \textit{Column Generator} stage, the property \textit{Options/Column to Generate} is only valid when \textit{Options/Column Method} is set to \textit{Explicit}.  Finally, we perform stage-specific or external consistency checks. For example, certain properties must match real-world artifacts—such as actual database connection names mentioned by the user. If such constraints are violated, the property is removed.

The results in Table~\ref{tab:property_prediction} show strong performance across all models, including the smallest LLaMA model.  This is partly due to our preprocessing of the utterance into focused sub-utterances, each matched to a corresponding list of possible properties.  This reduces the model’s job to a pattern-matching setup that even smaller models can achieve with high precision and competitive F1 scores.

\subsection{Pipeline Summary}
\label{sec:pipeline}

Our overall pipeline is shown in Figure~\ref{fig:architecture}. The process begins with a natural language utterance, which is split into sub-utterances and passed through a classification model to identify candidate stages. In parallel, a keyword matcher recovers additional stage candidates. Together, these form the retrieval step in our Classifier-Augmented Generation (CAG) approach. Based on the selected stages, we generate targeted descriptions and few-shot examples, which are then used by the LLM to predict the final list of stages. This stage prediction step is followed by a verification pass to ensure stage-level consistency.

Once the stages are known, the pipeline branches into two parallel steps: generating the nonlinear structure of the workflow (i.e., edge prediction), and inferring per-stage parameters (i.e., property prediction). For both tasks, the model uses the previously derived sub-utterances and stage descriptions. A final validation layer enforces edge cardinality constraints and property-level correctness. The result is a complete DataStage workflow generated from natural language input.

\section{Related Work}

Apart from our approach, we found Zap builder~\cite{Zapier2023-ai-powered} and Power Automate~\cite{Microsoft-power-automate} that both offer an AI-powered flow generation but no technical details or results about these tools seem to be available in the literature.  \cite{Datacamp2023-list} maintains a list of popular ETL tools but after closer inspection, none of them seem to support providing flows by a natural language description.

In the research literature, we found GOFA~\cite{Brachman2022-goal-driven} that creates application integration workflows through natural language. It utilizes an Integration Knowledge Graph, eliminating the need for an annotator. However, it serves mostly as generating a skeleton for a linear flow since it does not help with the configuration of the nodes' properties whereas our approach generates complete workflows.  FlowMind~\cite{zeng2023-flowmind} and AutoFlow~\cite{li2024-autoflow} are tools for executing a workflow based on a user’s query.  However, these ``workflows'' are executed ad hoc whereas our tool generates the implementation of a generic workflow that, once created, can be used repeatedly even without an LLM.  Analyza~\cite{Kedar2017-analyza} is a tool that generates SQL queries from natural language input, utilizing a parser, annotator, and table identifier to handle user queries. The tool addresses the challenge of ambiguity in complex SQL queries by initially limiting itself to simpler ones. Subsequent steps involve refining the original query using simple statements, akin to part of our approach of supporting the selection of a single ETL stage at a time. However, this work only describes the system and does not provide any evaluation and benchmarking.



\section{Limitations}

While our classifier achieves high accuracy, it was trained solely on single-label (utterance, stage) pairs due to the lack of sufficiently large multi-label training data. This constraint is mitigated by a final LLM-based multi-label generation step, but classification errors may still propagate when relevant candidate stages are excluded entirely.

Stage and property prediction remain robust across a range of workflows. However, edge prediction remains a key limitation. Although we achieve 73\% structural similarity to gold-standard layouts, exact edge match is only 37\%. As a result, the system often produces structurally valid drafts that still require user revision. In future work, we aim to explore hybrid architectures that combine LLM-based semantic reasoning with geometric deep learning methods such as graph neural networks (GNNs) to improve edge layout accuracy and flow topology prediction.

Our current validation logic assumes that table and column names are either correct or ignorable. While unmatched names are filtered during post-processing, the system does not currently attempt fuzzy matching or correction. This limits robustness in realistic settings where users may reference unavailable or misspelled schema elements.

Prompt formats and examples are tuned per model family, which may reduce portability across model architectures or providers. Generalization to other ETL platforms depends on the availability of operator-level metadata (e.g., transformation descriptions and property schemas), which is common but not universal.

Finally, while the system performs well on multi-step utterances, the style and structure of the inputs reflect internally sourced usage data. We expect further refinement to be needed when adapting to workflows from other domains or user populations.

\section{Conclusions}

We introduced a modular system for translating natural language into executable ETL workflows, combining classification and generation in a Classifier-Augmented Generation (CAG) architecture. The system decomposes user utterances into manageable sub-tasks—stage selection, edge prediction, and property configuration—paired with targeted prompting and validation.  CAG predicts correct workflow stages in over 97\% of cases, outperforms strong single-prompt and agentic baselines, and reduces token usage by more than 60\%. When deployed with a smaller model, it achieves comparable accuracy to the single-prompt baseline at roughly 90\% lower overall cost.  Property prediction achieves 90\% accuracy, and full-flow generation reaches over 70\% structural similarity—often requiring only minor corrections.  Beyond strong results, our architecture enables modular validation and constraint-based correction, and lends itself to transparent reasoning about intermediate predictions.  Edge prediction though remains a challenge but offers clear opportunities for optimization.  We believe our approach offers a promising foundation for LLM-assisted authoring of structured automation tasks in ETL and related domains.  Its successful integration into a production ETL platform further validates the practicality and scalability of the proposed architecture.

\bibliography{bibliography}


\appendix

\section{Example Utterances and Flows}

This appendix lists two examples prompts that are similar to the fewshot examples we used as well as examples for the evaluation.  Originally, this work started by predicting the stages used in linear flows and in a latest prototype was extended to cover non-linear flows as well.

\subsection{Linear Flow}

Linear flows do not employ any kind of branching and simply describes a sequence of actions the properties to be used for each action.  One such flow is the following:
\begin{quote}
I want to use teradata where my connection name is teradata-00, schema name is TM DS DB 1 and table name is EM- PLOYEE2. then sort on the age column. then filter out pizza column. then postgres where my connection name is tristan postconn , schema name is public and table name is demoautotest, Also do the following, Decimal rounding mode is ceiling, Generate Unicode Columns, Row limit should be 50.    
\end{quote}

For this natural language descriptionm, the following flow should be generated:

\begin{quote}
    \emph{teradata}
    $\rightarrow$
    \emph{filter}
    $\rightarrow$
    \emph{decode}
    $\rightarrow$
    \emph{column\_generator}
    $\rightarrow$
    \emph{postgresql}
\end{quote}

\subsection{Non-Linear Flows}

\begin{figure*}[ht]
    \centering
    \includegraphics[width=0.75\linewidth]{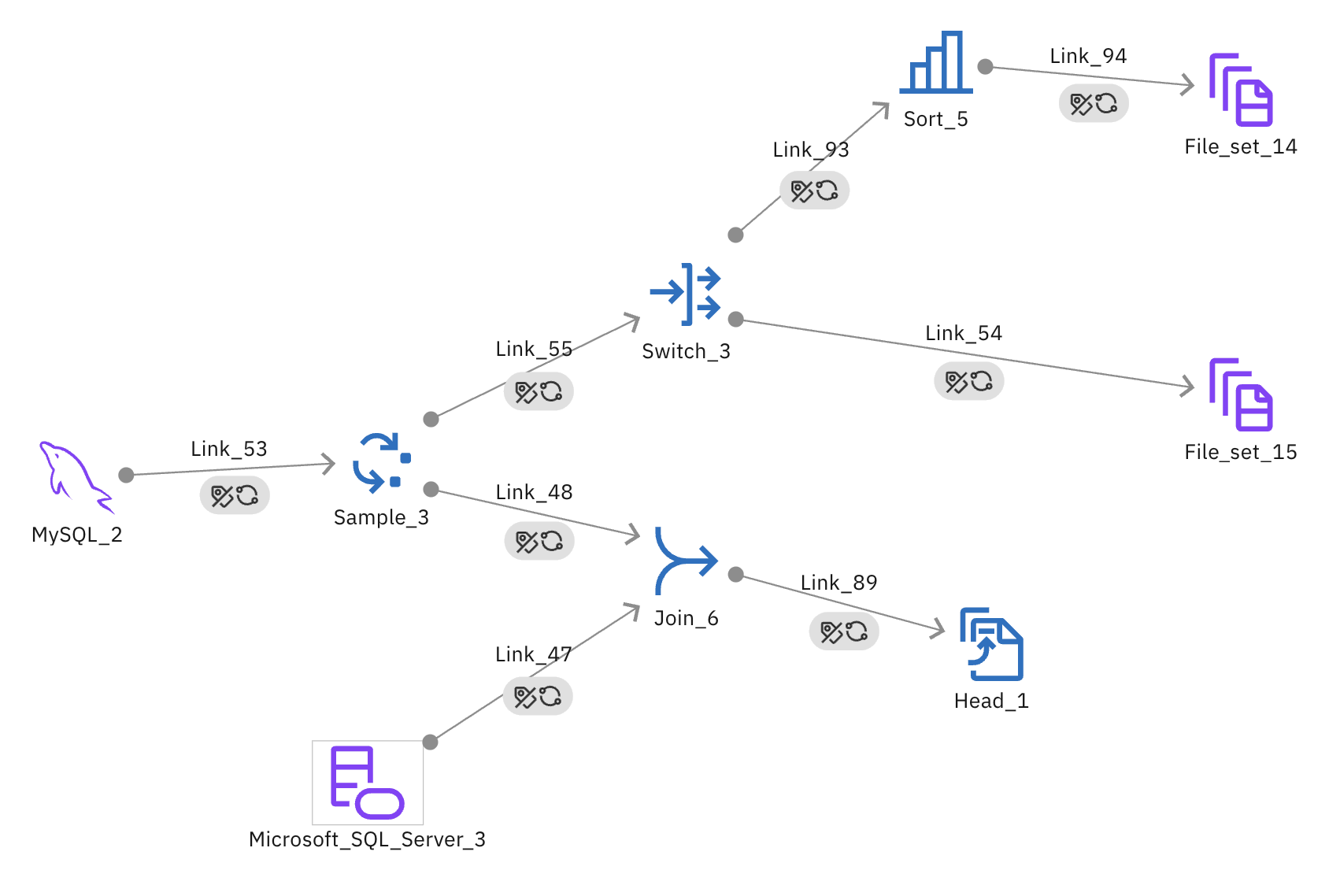}
    \caption{Non-Linear Flow Example}
    \label{fig:non-linear}
\end{figure*}

To evaluate the edge prediction to evaluate non-linear flows as well we used samples similar to the following which is shown grapically in Figure~\ref{fig:non-linear}:

\begin{quote}
    Extract data from MySQL and sample it using percent mode to send some data to a switch operator and the other data to a join operator. The switch stage writes some data to a fileset and outputs the rest to a sort stage that finally writes data into another fileset. The join operator merges the sampled MySQL data with data from a SQL Server source. Finally, the first few rows are selected using a head operator.
\end{quote}

\section{Prompt Templates}

Different model families may require tailored prompt formats for optimal prediction. In this work, we focused on the Granite and LLaMA model families and optimized their respective prompts for stage prediction. The following examples show the exact prompts used in the CAG setup for both Granite and LLaMA. By instead listing all stages and adjusting the few-shot examples accordingly, we obtain the prompt used in the single-prompt baseline.

\subsection{Granite Prompt Template}

Granite models require explicit role tokens to delimit system instructions, user input, and assistant responses. The prompt includes multiple few-shot examples and a final assistant cue that signals the model to complete the task. Its structure follows the public Granite documentation~\cite{granite2024docs}. 

The same base prompt structure is used for both the single-prompt and CAG setups: the only difference is that the single-prompt baseline enumerates all DataStage stages, whereas the CAG variant restricts this list to candidates identified by the classifier and keyword search (see Sections~\ref{sec:single} and~\ref{sec:cag}).

In this example, the operator description for \texttt{column\_generator} could not be automatically extracted by our tooling and was therefore omitted. Nevertheless, the LLM correctly inferred its semantics from the few-shot examples.

\begin{lstlisting}[style=prompt, caption={Granite prompt}, label={lst:pgranite}]
<|start_of_role|>system<|end_of_role|>Knowledge Cutoff Date: April 2024.
Today's Date: December 17, 2024.
You are Granite, developed by IBM. You are a helpful AI assistant.<|end_of_text|><|start_of_role|>user<|end_of_role|>Given the Context in the form of Operator and its descriptions, assign the correct Operator to the Utterance. Operators may occur multiple times. Only pick Operators given in the Context.
Your response should only include the answer. Do not provide any further explanation.

Context:
"column_generator": CREATE DESCRIPTION
"column_import": Column Import is a stage that imports data from a single column and outputs it to one or more columns
"dataset": A file datasource stage that reads and writes data from a DataSet/Data Set.
"dv": A stage that integrates data sources across multiple types and locations and turns all this data into one logical data view.
"head": The Head Stage selects the first N rows from each partition of an input data set and copies the selected rows to an output data set. You can sample data using this stage
"split_subrecord": The Split Subrecord stage separates an input subrecord field into a set of top-level vector columns.
"split_vector": The Split Vector operator stage modifies an input vector column by splitting it into columns
"tail": The tail operator copies the last N records from each partition of its input data set to its output data set. By default, N is 10 records

Here are some examples, complete the last one:
Utterance: Use Data Set. Please take action to Ignore missing columns. 
Operators: "dataset"

Utterance: Use split vector to split input vector 'Category'
Operators: "split_vector"

Utterance: Starting from row 4, get 100 rows from my dataset
Operators: "head"

Utterance: Retrieve data from Dropbox and process it using Data Virtualization. Then, utilize Data Virtualization Manager to provide read and write access to IBM Z data in place.
Operators: "dropbox, dv, dvm"

Utterance: I want to use column generator to make the column 'ID'. Please don't combine underlying operators into a single process
Operators: "column_generator"

Utterance: Read data from Azure File Storage and process the tail of the data, which includes the last N records from each partition. Finally, store the processed data in Azure Blob Storage.
Operators: "azure_file_storage, tail, azure_blob_storage"

Utterance: I want to use column import to import a column to the output column 'Region'. Don't keep the input column and don't combine operators.
Operators: "column_import"

Utterance: Separate subrecord field column_1 into a set of top-level vector columns.
Operators: "split_subrecord"

Utterance: Extract data from dataset temp.ds and ignore if there are any missing columns
Operators: "dataset"

Utterance: head
Operators: "head"

Utterance: Use the stage called dv. Use the connection with the GUID 'connectionGUID', set the Java heap size to 256 MegaBytes, and round to the nearest even number. Set the default maximum length for column names to 20000 characters, use the general mode to read records from the table, and return a maximum of 10 rows. I want to read from the schema named 'GOSALESHR_1021 and read data from the table named 'EMPLOYEE_EXPENSE_DETAIL'. 
Operators: "dv"

Utterance: I want to split the vector column 'Data'. Also, don't use underlying operators.
Operators: "split_vector"

Utterance: Apply encoded change operations to a before data set based on a changed data set. Then, generate new columns for the dataset. Finally, store the modified dataset in IBM Cloud Object Storage.
Operators: "change_apply, column_generator, cloud_object_storage"

Utterance: Import a column to the columns 'Name', 'Age', 'Region', 'Gender', 'Department', 'ID', 'Salary'. Auto combine operators
Operators: "column_import"

Utterance: tail
Operators: "tail"

Utterance: Extract records where sales exceed $1000, split the address field into street, city, and country, and then integrate this data with customer information using department_id.
Operators: "filter, split_subrecord, join"

Utterance: Select the first and last two records of the dataset and then print their details for quality check purposes.
Operators: "head, tail, peek"

Utterance: Remove all but the first 10 rows of data from my dataset
Operators: "head"

Utterance: Write to a dataset
Operators: "dataset"

Utterance: I want to read from ibm data virtualization
Operators: "dv"

Utterance: Use column generator to create the column region
Operators: "column_generator"

Utterance: Split the vector column 'Name'
Operators: "split_vector"

Utterance: Use column import
Operators: "column_import"

Utterance: Use Tail
Operators: "tail"

Utterance: Split subrecord parent to a set of similarly named and typed top-level columns.
Operators: "split_subrecord"

Utterance: Split the input subrecord field into a set of top-level vector columns using the Split Subrecord stage. Then, modify the input vector column by splitting it into columns using the Split Vector stage.
Operators: "split_subrecord, split_vector"

Utterance: Read data from the Data Set file 'test.ds' and if there are any missing columns, set them to the null value.
Operators: "dataset"

Utterance: Retrieve data from Dropbox and integrate it using Data Virtualization to create a unified data view.
Operators: "dropbox, dv"

Utterance: Return rows of data with a period of 5.
Operators: "head"

Utterance: split vector
Operators: "split_vector"

Utterance: CREATE DESCRIPTION
Operators: "column_generator"

Utterance: Using tail operator as an intermediate stage, copy the last 5 records, from partitions 1,2 and 3 of input dataset to output data set
Operators: "tail"

Utterance: Combine data from multiple input datasets such as sales_data, customer_info, and product_catalog into a single output dataset for comprehensive analysis; Sorted on ID in ascending order. then divide the address field into separate columns for street, city, and pincode.
Operators: "funnel, split_subrecord"

Utterance: Import the column 'Info' to output column 'ID'. If there are failing rows, fail the stage, and don't combine operators. I don't want to keep the import column.
Operators: "column_import"

Utterance: Overwrite rows from data set 'test.ds'
Operators: "dataset"

Utterance: Read from IBM data virtualization
Operators: "dv"

Utterance: Use the stage called Head. Please combine operators into a single process and set the period to 5 for row selection. 
Operators: "head"

Utterance: Generator columns with schema file 'employee.json'. I want to combine underlying operators into a single process
Operators: "column_generator"

Utterance: Split the vector column 'Quantity' and combine underlying operators
Operators: "split_vector"

Utterance: I want to separate an input subrecord field parent into a set of top-level vector columns.
Operators: "split_subrecord"

Utterance: Import columns with schema file 'sales.xml'. Log failing rows and combine underlying operators.
Operators: "column_import"

Utterance: Select the first and last five records of the Employee dataset and select the first ten records from Student dataset.
Operators: "head, tail, head"

Utterance: Using dataset, overwrite rows from the data set named test.ds
Operators: "dataset"

Utterance: Get all rows except the first 10 rows from all partitions.
Operators: "head"

Utterance: I want to read table EconomicOutput from IBM Data Virtualization, use general read method, apply schema WorldEconomicForum, use GDP as the key column, round decimals to the ceiling, generate unicode for columns.
Operators: "dv"

Utterance: Use the file 'employee.json' to generate a column. Please combine underlying operators
Operators: "column_generator"

Utterance: Process complex flat files, split vector columns, and load the data into a SingleStore Database.
Operators: "complex_flat_file, split_vector, singlestore"

Utterance: Give me the last 3 rows of my input dataset
Operators: "tail"

Utterance: The Split Subrecord stage separates an input subrecord field into a set of top-level vector columns.
Operators: "split_subrecord"

Utterance: Import from column 'Sales' using file 'sales_schema.xml'. Keep the import column
Operators: "column_import"

Utterance:
Split the full_name field of the employee_data dataset into separate columns for first_name and last_name, then capitalize the first letter of each name for consistency.
Operators:<|end_of_text|><|start_of_role|>assistant<|end_of_role|>
\end{lstlisting}

\subsection{Llama Prompt Template}

The LLaMA prompt follows a similar structure but uses instruction-tuned phrasing and omits explicit role tokens. Unlike Granite, LLaMA models do not require these delimiters to correctly complete the final example. However, it is important to pre-seed the output with the opening quote character (\texttt{"}) to guide the model toward the correct output format; without this cue, the LLaMA models occasionally deviate from the expected format. Interestingly, for the Granite models this pre-seeding had the opposite effect and reduced the output consistency.

\begin{lstlisting}[style=prompt, caption={Llama prompt}, label={lst:pllama}]
Given the Context in the form of Operator and its descriptions, assign the correct Operator to the Utterance. Operators may occur multiple times. Only pick Operators given in the Context.
Your response should only include the answer. Do not provide any further explanation.

Context:
"column_generator": CREATE DESCRIPTION
"column_import": Column Import is a stage that imports data from a single column and outputs it to one or more columns
"dataset": A file datasource stage that reads and writes data from a DataSet/Data Set.
"dv": A stage that integrates data sources across multiple types and locations and turns all this data into one logical data view.
"head": The Head Stage selects the first N rows from each partition of an input data set and copies the selected rows to an output data set. You can sample data using this stage
"split_subrecord": The Split Subrecord stage separates an input subrecord field into a set of top-level vector columns.
"split_vector": The Split Vector operator stage modifies an input vector column by splitting it into columns
"tail": The tail operator copies the last N records from each partition of its input data set to its output data set. By default, N is 10 records

Here are some examples, complete the last one:
Utterance: Use Data Set. Please take action to Ignore missing columns. 
Operators: "dataset"

Utterance: Use split vector to split input vector 'Category'
Operators: "split_vector"

Utterance: Starting from row 4, get 100 rows from my dataset
Operators: "head"

Utterance: Retrieve data from Dropbox and process it using Data Virtualization. Then, utilize Data Virtualization Manager to provide read and write access to IBM Z data in place.
Operators: "dropbox, dv, dvm"

Utterance: I want to use column generator to make the column 'ID'. Please don't combine underlying operators into a single process
Operators: "column_generator"

Utterance: Read data from Azure File Storage and process the tail of the data, which includes the last N records from each partition. Finally, store the processed data in Azure Blob Storage.
Operators: "azure_file_storage, tail, azure_blob_storage"

Utterance: I want to use column import to import a column to the output column 'Region'. Don't keep the input column and don't combine operators.
Operators: "column_import"

Utterance: Separate subrecord field column_1 into a set of top-level vector columns.
Operators: "split_subrecord"

Utterance: Extract data from dataset temp.ds and ignore if there are any missing columns
Operators: "dataset"

Utterance: head
Operators: "head"

Utterance: Use the stage called dv. Use the connection with the GUID 'connectionGUID', set the Java heap size to 256 MegaBytes, and round to the nearest even number. Set the default maximum length for column names to 20000 characters, use the general mode to read records from the table, and return a maximum of 10 rows. I want to read from the schema named 'GOSALESHR_1021 and read data from the table named 'EMPLOYEE_EXPENSE_DETAIL'. 
Operators: "dv"

Utterance: I want to split the vector column 'Data'. Also, don't use underlying operators.
Operators: "split_vector"

Utterance: Apply encoded change operations to a before data set based on a changed data set. Then, generate new columns for the dataset. Finally, store the modified dataset in IBM Cloud Object Storage.
Operators: "change_apply, column_generator, cloud_object_storage"

Utterance: Import a column to the columns 'Name', 'Age', 'Region', 'Gender', 'Department', 'ID', 'Salary'. Auto combine operators
Operators: "column_import"

Utterance: tail
Operators: "tail"

Utterance: Extract records where sales exceed $1000, split the address field into street, city, and country, and then integrate this data with customer information using department_id.
Operators: "filter, split_subrecord, join"

Utterance: Select the first and last two records of the dataset and then print their details for quality check purposes.
Operators: "head, tail, peek"

Utterance: Remove all but the first 10 rows of data from my dataset
Operators: "head"

Utterance: Write to a dataset
Operators: "dataset"

Utterance: I want to read from ibm data virtualization
Operators: "dv"

Utterance: Use column generator to create the column region
Operators: "column_generator"

Utterance: Split the vector column 'Name'
Operators: "split_vector"

Utterance: Use column import
Operators: "column_import"

Utterance: Use Tail
Operators: "tail"

Utterance: Split subrecord parent to a set of similarly named and typed top-level columns.
Operators: "split_subrecord"

Utterance: Split the input subrecord field into a set of top-level vector columns using the Split Subrecord stage. Then, modify the input vector column by splitting it into columns using the Split Vector stage.
Operators: "split_subrecord, split_vector"

Utterance: Read data from the Data Set file 'test.ds' and if there are any missing columns, set them to the null value.
Operators: "dataset"

Utterance: Retrieve data from Dropbox and integrate it using Data Virtualization to create a unified data view.
Operators: "dropbox, dv"

Utterance: Return rows of data with a period of 5.
Operators: "head"

Utterance: split vector
Operators: "split_vector"

Utterance: CREATE DESCRIPTION
Operators: "column_generator"

Utterance: Using tail operator as an intermediate stage, copy the last 5 records, from partitions 1,2 and 3 of input dataset to output data set
Operators: "tail"

Utterance: Combine data from multiple input datasets such as sales_data, customer_info, and product_catalog into a single output dataset for comprehensive analysis; Sorted on ID in ascending order. then divide the address field into separate columns for street, city, and pincode.
Operators: "funnel, split_subrecord"

Utterance: Import the column 'Info' to output column 'ID'. If there are failing rows, fail the stage, and don't combine operators. I don't want to keep the import column.
Operators: "column_import"

Utterance: Overwrite rows from data set 'test.ds'
Operators: "dataset"

Utterance: Read from IBM data virtualization
Operators: "dv"

Utterance: Use the stage called Head. Please combine operators into a single process and set the period to 5 for row selection. 
Operators: "head"

Utterance: Generator columns with schema file 'employee.json'. I want to combine underlying operators into a single process
Operators: "column_generator"

Utterance: Split the vector column 'Quantity' and combine underlying operators
Operators: "split_vector"

Utterance: I want to separate an input subrecord field parent into a set of top-level vector columns.
Operators: "split_subrecord"

Utterance: Import columns with schema file 'sales.xml'. Log failing rows and combine underlying operators.
Operators: "column_import"

Utterance: Select the first and last five records of the Employee dataset and select the first ten records from Student dataset.
Operators: "head, tail, head"

Utterance: Using dataset, overwrite rows from the data set named test.ds
Operators: "dataset"

Utterance: Get all rows except the first 10 rows from all partitions.
Operators: "head"

Utterance: I want to read table EconomicOutput from IBM Data Virtualization, use general read method, apply schema WorldEconomicForum, use GDP as the key column, round decimals to the ceiling, generate unicode for columns.
Operators: "dv"

Utterance: Use the file 'employee.json' to generate a column. Please combine underlying operators
Operators: "column_generator"

Utterance: Process complex flat files, split vector columns, and load the data into a SingleStore Database.
Operators: "complex_flat_file, split_vector, singlestore"

Utterance: Give me the last 3 rows of my input dataset
Operators: "tail"

Utterance: The Split Subrecord stage separates an input subrecord field into a set of top-level vector columns.
Operators: "split_subrecord"

Utterance: Import from column 'Sales' using file 'sales_schema.xml'. Keep the import column
Operators: "column_import"

Utterance: Split the full_name field of the employee_data dataset into separate columns for first_name and last_name, then capitalize the first letter of each name for consistency.
Operators: "
\end{lstlisting}

\end{document}